%% file: main.tex
\documentclass[11pt, a4paper, logo, twocolumn, external, copyright, nonumbering]{googledeepmind}

\pdftrailerid{redacted}

\makeatletter
\renewcommand\bibentry[1]{\nocite{#1}{\frenchspacing\@nameuse{BR@r@#1\@extra@b@citeb}}}
\makeatother

\usepackage{kantlipsum, lipsum}
\usepackage{dsfont}
\usepackage{gdm-colors}
\usepackage{booktabs}
\usepackage{lscape}

\usepackage{algorithm2e}

\usepackage[authoryear, sort&compress, round]{natbib}

\graphicspath{{figures/}}

\title{Model Predictive Simulation Using Structured Graphical Models and Transformers}


\reportnumber{001} 

\renewcommand{\today}{2024-06-11}

\newcommand{\eat}[1]{} 

\author[1]{Xinghua Lou}
\author[1]{Meet Dave}
\author[1]{Shrinu Kushagra}
\author[1]{Miguel L\'azaro-Gredilla}
\author[1]{Kevin Murphy}

\affil[1]{Google DeepMind}

\input{abstract}

\begin{document}

\maketitle

\input{intro}

\input{method}

\input{eval}
\input{conclusion}

\setlength{\bibsep}{3pt}
\bibliographystyle{abbrvnat}
\nobibliography*
\scriptsize{\bibliography{refs}}

\input{appendix}

\end{document}

%% file: abstract.tex
\begin{abstract}
We propose an approach to simulating trajectories of multiple interacting agents (road users)
based on transformers
and probabilistic graphical models (PGMs),
and apply it to the Waymo SimAgents challenge.
The transformer baseline is based on the
MTR model \citep{shi2024mtr++},
which predicts multiple future trajectories
conditioned on the past trajectories
and static road layout features.
We then improve upon these generated trajectories
using a PGM, which contains  factors which encode prior knowledge, such as a preference for smooth trajectories, and avoidance of collisions with static obstacles and other moving agents.
We perform (approximate) MAP inference in this PGM using the Gauss-Newton method.
Finally we sample $K=32$ trajectories
for each of the $N \sim 100$ agents for the next
$T=8 \Delta$ time steps,
where $\Delta=10$ is the sampling rate per second.
Following the Model Predictive Control (MPC) paradigm,
we only return the first element of our forecasted
trajectories at each step,
and then we replan, so that the simulation can constantly adapt to its changing environment.
We therefore call our approach "Model Predictive Simulation" or MPS.
We show that MPS improves upon the MTR baseline,
especially in safety critical metrics such as collision rate.
Furthermore, our approach is compatible with any underlying forecasting model, and does not require extra training, so we believe it is a valuable contribution to the community.
\end{abstract}

%% file: intro.tex
\section{Introduction}

The use of transformers to create  generative models to simulate agent trajectories,
trained on large datasets such as Waymo Open Data
 \citep{ettinger2021large},
 has become very popular in recent years.
Most previous work has been focusing on improving the architecture \citep{nayakanti2023wayformer,shi2024mtr++}, 
the training objective \citep{ngiam2021scene,shi2024mtr++},
the trajectory representation \citep{seff2023motionlm,philion2023trajeglish} or the speed \citep{zhou2023query} of these transformer-based models.

This paper tackles the problem from an orthogonal and complementary angle -- namely the use of prior knowledge, encoded using a probabilistic graphical model (PGM).
We perform approximate MAP inference in the PGM to ``post process''
the trajectory proposals from a base transformer
model, to increase their realism
and compliance with constraints, such as collision avoidance.

To ensure that our predicted forecasts are adaptive to the changing environment, we replan at each step, following the principle of  model predictive control (MPC),
which is widely used for controlling complex dynamical systems \citep{schwenzer2021review}.
We therefore call our approach 
\textbf{Model Predictive Simulation (MPS)}.

Our MPS approach differs from previous PGM methods for trajectory simulation, such as 
 JFP \citep{luo2023jfp},
 in several ways.
 First, we explicitly  include (data-dependent) factors
 for collision avoidance
 and smooth trajectories,
 so we have better control over the generated trajectories. 
 Second, our approach is iterative (being based on MPC), while JFP commits to the trajectory proposals at $t=0$ and is thus open loop.
 Third, our approach uses the Gauss-Newton method to compute the joint MAP estimate,
 whereas JFP is based on discrete belief propagation methods to choose amongst a finite
 set of candidate trajectories.

%% file: method.tex
\section{Method}

\newcommand{\context}{\ensuremath{\mathbf{c}}}
\newcommand{\history}{\ensuremath{\mathbf{h}_{1:N}}}

\paragraph{Outer loop}

\RestyleAlgo{ruled}
\SetAlgoNoLine
\begin{algorithm}
\footnotesize
\caption{SimAgents outer loop}
\label{alg:SimAgents}

\KwIn{
Scene context \context, 
num. agents $N$,
num. samples $K$,
trajetory length $T$}

\KwOut{Sampled trajectories,
$h_{1:N}^{1:K,1:T}$}

\For{$k = 1$ to $K$}{
    $s_{1:N}^{k,0} = \text{init-trajectory}(\context)$

    \For{$t=1$ to $T$}{

        \text{Sample} $r_{1:N}^{k,t} = \text{MPS}(\context, s_{1:N}^{k, 1:t-1})$

        \text{Extend} $s_{1:N}^{k,1:t} = \text{append}(s_{1:N}^{k,1:t}, r_{1:N}^{k,t})$
    }
}

\end{algorithm}

The overall simulation pseudocode is shown in Algo.~\ref{alg:SimAgents}.
It generates a set of $K=32$ trajectories,
each of length $T=80$,
for $N$ agents
given the scene context \context.
(The exact value of $N$ depends
on  the number of agents that
are visible in
$\mathbf{c}$.)
We denote the generated output by
$s_{1:N}^{1:K,1:T}$,
where $s_i^{k,t} = [x,y,\dot{x},\dot{y}]$ is the state (2d location and velocity) of the $i$'th agent
in sample $k$.

\paragraph{Inner loop}

\begin{algorithm}
\caption{Model Predictive Simulation}
\label{alg:mps}
\footnotesize

\KwIn{
Scene context $\mathbf{c}$, 
agent history $h_{1:N}$, 
num. agents $N$,
future planning horizon $F$,
number of rollouts $J$,
transformer proposal $\pi$
}

\KwOut{Predicted next state for each agent,
$r_{1:N}$}

\For{$j = 1$ to $J$} {
    \text{Sample} $(a_{1:N}^{j,1:F}, g_{1:N}^j) \sim \pi( \mathbf{c},  h_{1:N}, F)$

    $G^j = \text{BuildFactorGraph}(a_{1:N}^{j,1:F}, g_{1:N}^j, \context)$

    \text{Initialize} $s_{1:N}^{j,1:F} = a_{1:N}^{j,1:F}$

    $(s_{1:N}^{j,1:F}, E^j) = 
    \text{Inference}(G^j, s_{1:N}^{j,1:F})$  

}

\text{Sample} $j^* \sim  \text{SoftMin}(E^{1:J})$

\Return $s_{1:N}^{j^*,1}$

\end{algorithm}

At each step $t$, the simulator calls our MPS
algorithm to generate a prediction
for  the next state of each agent. The pseudocode for this is shown in Algo.~\ref{alg:mps}.
The approach is as follows.
First we use the MTR transformer model $\pi$
\citep{shi2024mtr++}
to sample a set of $N$  goal locations,
$g_{1:N}^j$, one for each agent,
as well as a sequence of anchor points
leading to each goal,
$a_{1:N}^{j,1:F}$,
where $F$ is the planning or forecast horizon.
We do this $J=60$ times in parallel,
to create a set of possible futures.
We then use the PGM to generate $J$
joint trajectories (for all $N$ agents), using the method described below.
Finally we evaluate the energy of each generated
trajectory, $E^j$, 
sample one of the low energy (high probability) ones to get $s_{1:N}^{j^*,1:F}$,
and return the first step of this sampled trajectory,
$s_{1:N}^{j^*,1}$.

\paragraph{Graphical model}

\begin{figure}[htb]
	\centering
	\includegraphics[width=\columnwidth]{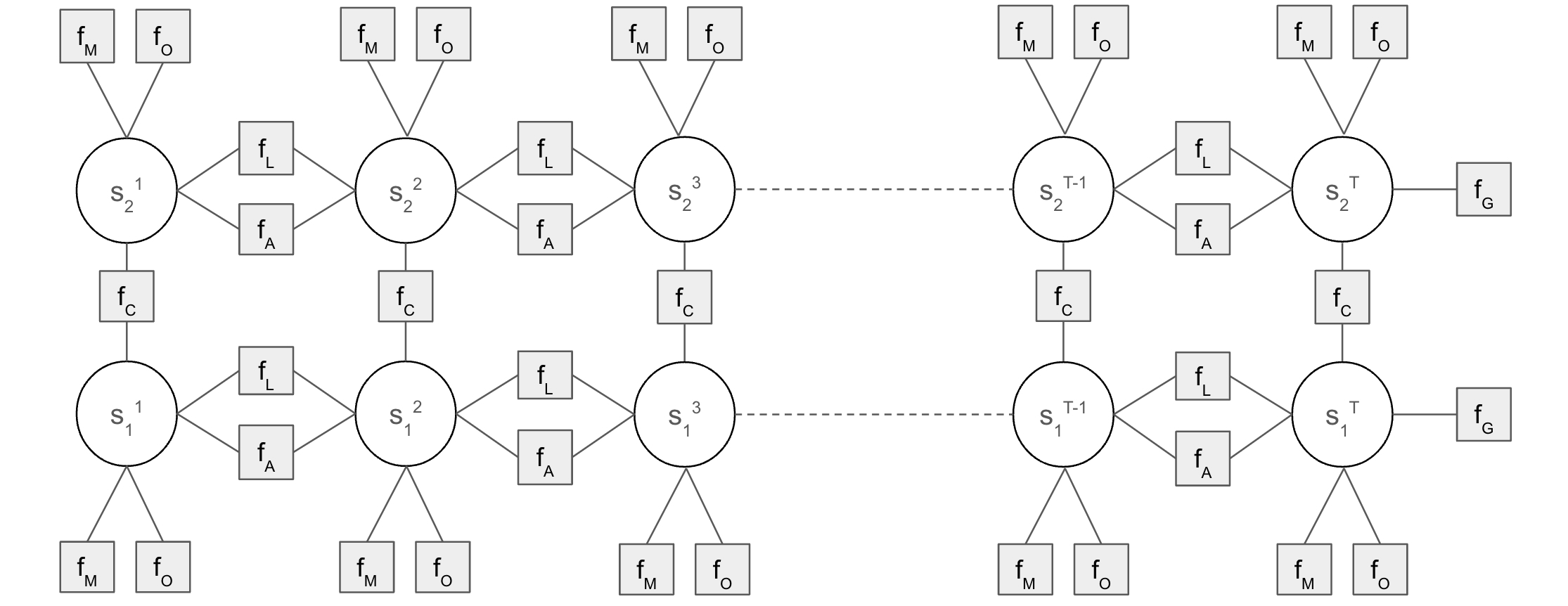}
	\caption{Factor Graph for $N=2$ agents
	unrolled for $T$ planning steps.
	Circles are random variables,
	gray squares are fixed factors.
	}
	\label{fig:factor-graph}
\end{figure}

\begin{figure}[htb]
    \centering
    \footnotesize
\begin{align*}
p(\mathbf{s}^{1:F}_{1..N} \mid \context, 
\mathbf{a}_{1:N}^{1:F},g_{1:N}) \propto
&
\prod_{i=1}^N  \left[ 
 f_G(s_i^F | g_i)
 \cdot
  \prod_{t=1}^{F-1}
 f_M(s_i^{t} |a_i^{t})
\right] \cdot
\\
&
\prod_{i=1}^N  \left[
\prod_{t=2}^F
f_L(s_i^{t-1}, s_i^{t}) 
\cdot f_A(s_i^{t-1}, s_i^{t}) 
\right] \cdot
\\
& 
\prod_{i=1}^N  \prod_{t=1}^F 
f_O(s_i^{t} \mid \mathbf{c})
\cdot
\prod_{i \ne j} \prod_{t=1}^F
f_C(s_i^{t}, s_j^{t})
\label{eq:PGM}
\end{align*}
    \caption{Joint probability model.}
    \label{fig:joint}
\end{figure}

The key to our method is the probabilistic graphical model (PGM) for improving upon the proposed trajectories by MTR.
The factor graph is shown 
in Fig.~\ref{fig:factor-graph}
and the corresponding conditional joint distribution is given in Fig.~\ref{fig:joint}. The model was inspired by \citep{Patwardhan2022} who uses Gaussian belief propagation.
We now explain each of the factors.

First we have factors which compare
a candidate trajectory to the original proposal.
The motion factor is defined as 
$f_\mathrm{M}(\mathbf{s}_i^{t}) = |s_i^t - a_i^t|$,
where $a_i^t$ is the predicted location (anchor point)
for agent $i$ at time $t$ as computed by $\pi$.
This ensures the trajectory stays close to the initial proposal.
The proximity to goal factor is defined as 
$f_\mathrm{G}(s_i^F) = |s_i^F - g_i|$,
where  $g_i$ is the goal for agent $i$
predicted  by $\pi$.
This ensures the trajectory ends close to where we expect.

Second we have factors defined from "physics".
We define a factor that penalizes
deviation from linear motion:
$f_\mathrm{L}(s, s') = |
s'_{x y} - (s_{x y} + s_{\dot{x} \dot{y}} \Delta_t) |$,
where $s_{xy}$ are the location components of $s$,
$s_{\dot{x}\dot{y}}$ are the velocity components of $s$,
and $\Delta t$ is the sampling rate.
We also define a factor that penalizes
change in direction:
$f_\mathrm{A}(s, s') = | s_{\dot{x}\dot{y}} - s'_{\dot{x}\dot{y}} |$.
We used weight 2.0 for $f_\mathrm{A}$ and 1.0 for all other factors.

Third we have factors derived from static obstacles
on the road:
$f_\mathrm{O} (s \mid \mathbf{c}) = \max_{(x, y) \in \mathbf{c}_\mathrm{RE}}  \mathrm{G} (x, y \mid s)$,
where $\mathbf{c}_\mathrm{RE}$ represents the coordinates of the road edges (part of the context $\mathbf{c}$) and $\mathrm{G} (x, y \mid s)$ is a Gaussian field centered and rotated according to the agent's location $s_{xy}$.

Finally, we have pairwise collision factors between agents:
$f_\mathrm{C}(s, s') = \max_{(x, y) \in \mathrm{CCP}(s')}  \mathrm{G} (x, y \mid s)$,
where $\mathrm{G} (x, y \mid s)$ is a  Gaussian field for agent $s$, and $\mathrm{CCP}(s')$
are the 9 collision checking points (CCP) for the other agent $s'$ (4 corners, 4 centers of the sides, and center of the agent).

\paragraph{Inference}

Inference on the factor graph is equivalent to minimizing a non-linear, non-convex quadratic optimization problem defined over $\mathbf{s}_{1:N}^{1:F}$.
For efficiency reasons, we developed a two-step approach. First, we use the Gauss–Newton method 
to solve a partial model that only consists of $f_\mathrm{M}$, $f_\mathrm{L}$ and $f_\mathrm{A}$ factors,
as these can all be evaluated in parallel
across agents using $N$
individual trajectory models. 
This step produces smoothed trajectories, which are then frozen. 
Second, we sample joint trajectories for agents according to their probability (unnormalized energy),
and use the $f_\mathrm{O}$ and $f_\mathrm{C}$ factors
to score their quality. After repeating this $J$ times, the best joint trajectories are sampled from a softmin operation over the scores of the $J$ samples.

%% file: eval.tex
\begin{table*}[ht]
\centering
\resizebox{\textwidth}{!}{
\begin{tabular}{lccccccccccc}
\toprule
\textbf{WAYMO} & \multicolumn{1}{c}{\textbf{META METRIC}} & \multicolumn{4}{c}{\textbf{KINEMATIC}} & \multicolumn{3}{c}{\textbf{INTERACTIVE}} & \multicolumn{2}{c}{\textbf{MAP}} & \\
\cmidrule(lr){2-2} \cmidrule(lr){3-6} \cmidrule(lr){7-9} \cmidrule(lr){10-11}
\textbf{LEADERBOARD} & \textbf{REALISM} & \textbf{\begin{tabular}[c]{@{}c@{}}LINEAR \\ SPEED\end{tabular}} & \textbf{\begin{tabular}[c]{@{}c@{}}LINEAR \\ ACCEL.\end{tabular}} & \textbf{\begin{tabular}[c]{@{}c@{}}ANG. \\ SPEED\end{tabular}} & \textbf{\begin{tabular}[c]{@{}c@{}}ANG. \\ ACCEL.\end{tabular}} & \textbf{\begin{tabular}[c]{@{}c@{}}DIST. \\ TO OBJ.\end{tabular}} & \textbf{COLLISION} & \textbf{TTC} & \textbf{\begin{tabular}[c]{@{}c@{}}DIST. \\ TO ROAD\end{tabular}} & \textbf{OFFROAD} & \textbf{minADE $\downarrow$} \\
\midrule
SMART &
  \textbf{0.7511} &
  \textbf{0.3646} &
  \textbf{0.4057} &
  0.4231 &
  0.5844 &
  \textbf{0.3769} &
  \textbf{0.9655} &
  \textbf{0.8317} &
  0.6590 &
  0.9362 &
  1.5447 \\
MVTE &
  0.7301 &
  0.3506 &
  0.3530 &
  \textbf{0.4974} &
  \textbf{0.5999} &
  0.3742 &
  0.9049 &
  0.8309 &
  \textbf{0.6655} &
  0.9071 &
  1.6769 \\
MPS (Ours) &
  0.7416 &
  0.3137 &
  0.3049 &
  0.4705 &
  0.5834 &
  0.3593 &
  0.9629 &
  0.8070 &
  0.6651 &
  \textbf{0.9366} &
  \textbf{1.4841} \\

\bottomrule
\end{tabular}}
\caption{WOSAC Leaderboard: SMART (2024 winner) Vs. MVTE (2023 winner) Vs. MPS (ours).}
\label{tab:wosac_leaderboard}
\end{table*}

\section{Experimental Evaluation}

\noindent
\textbf{Benchmark} We evaluated MPS on the 2024 Waymo Open Sim Agents Challenge
\citep{wosac}, 
where the task is simulating 32 realistic rollouts of all agents in the scene given their 1s history for 8s into the future. The simulation needs to be closed-loop and factorized between the ADV and other agents, which MPS satisfies naturally.

\noindent
\textbf{Implementation Details} We implemented the factors and the inference in JAX
\footnote{\url{https://github.com/google/jax}}
and JAXopt
\footnote{\url{https://jaxopt.github.io/}}
for the Gauss-Newton method. We leveraged JAX's just-in-time (JIT) compilation and observed great scalability. For speed up, we take 10 immediate next steps at each MPS iteration.

We trained our own MTR model $\pi$
using the open source code
\footnote{\url{https://github.com/sshaoshuai/MTR}}
. We removed local attention and reduced the source polylines to 512. The training data
is augmented by adding extra interacting agents,
and by applying random history dropouts. We followed the original training setup except the number of epochs (50), the batch size (8) and the LR schedule ([25, 30, 35, 40, 45]). Training took about 3 days on 16 A100s.

We used only the official Waymo Open Motion Dataset v1.2.1
and did \textbf{not} use any Lidar or Camera data. We did \textbf{not} need any additional training and we did \textbf{not} use ensembles.

\noindent
\textbf{Sim Agents 2024 Results} We ranked number 4 among all methods (Table~\ref{tab:wosac_leaderboard}). We outperformed the 2023 winner MVTE \citep{wang2023multiverse} which also uses MTR \citep{shi2024mtr++}, and are approximately 1 point behind the 2024 winner SMART \citep{wu2024smart}. MPS achieved near-top performance in a few safety critical metrics such as COLLISION and OFFROAD, showing the effectiveness of the priors in our model. MPS showed a lack of performance in LINEAR SPEED / ACCEL. We speculate this is because MPS can generate diverse rollouts that are very different from the logged data used for metric evaluation.

\begin{table*}[htb]
\centering
\resizebox{\textwidth}{!}{
\begin{tabular}{lccccccccccc}
\toprule
\textbf{} & \multicolumn{1}{c}{\textbf{META METRIC}} & \multicolumn{4}{c}{\textbf{KINEMATIC}} & \multicolumn{3}{c}{\textbf{INTERACTIVE}} & \multicolumn{2}{c}{\textbf{MAP}} & \\
\cmidrule(lr){2-2} \cmidrule(lr){3-6} \cmidrule(lr){7-9} \cmidrule(lr){10-11}
\textbf{METHOD} & \textbf{REALISM} & \textbf{\begin{tabular}[c]{@{}c@{}}LINEAR \\ SPEED\end{tabular}} & \textbf{\begin{tabular}[c]{@{}c@{}}LINEAR \\ ACCEL.\end{tabular}} & \textbf{\begin{tabular}[c]{@{}c@{}}ANG. \\ SPEED\end{tabular}} & \textbf{\begin{tabular}[c]{@{}c@{}}ANG. \\ ACCEL.\end{tabular}} & \textbf{\begin{tabular}[c]{@{}c@{}}DIST. \\ TO OBJ.\end{tabular}} & \textbf{COLLISION} & \textbf{TTC} & \textbf{\begin{tabular}[c]{@{}c@{}}DIST. \\ TO ROAD\end{tabular}} & \textbf{OFFROAD} & \textbf{minADE $\downarrow$} \\
\midrule
MTR+RAND &
  0.7019 &
  \textbf{0.3922} &
  \textbf{0.3530} &
  0.3899 &
  0.3304 &
  \textbf{0.3691} &
  0.8491 &
  \textbf{0.8164} &
  \textbf{0.6706} &
  0.9207 &
  \textbf{1.3084} \\
MPS &
  \textbf{0.7418} &
  0.3158 &
  0.3056 &
  \textbf{0.4664} &
  \textbf{0.5818} &
  0.3604 &
  \textbf{0.9617} &
  0.8094 &
  0.6651 &
  \textbf{0.9374} &
  1.4841 \\

\bottomrule
\end{tabular}}
\caption{Abalation study -- comparing MPS to MTR with random trajectory sampling.}
\label{tab:mps-vs-mtr-rand}
\end{table*}

\noindent
\textbf{Ablation Study} To evaluate the value of the PGM priors, we compare MPS to the same MTR model with random trajectory sampling (MTR+RAND) on the validation dataset. As shown in Table~\ref{tab:mps-vs-mtr-rand}, MPS improved safety-critical metrics such as COLLISION, OFFROAD and the overall REALISM score, while lacked performance at LINEAR SPEED / ACCEL for the same reason discussed above.

\begin{figure}[htb]
	\centering
	\includegraphics[width=\columnwidth]{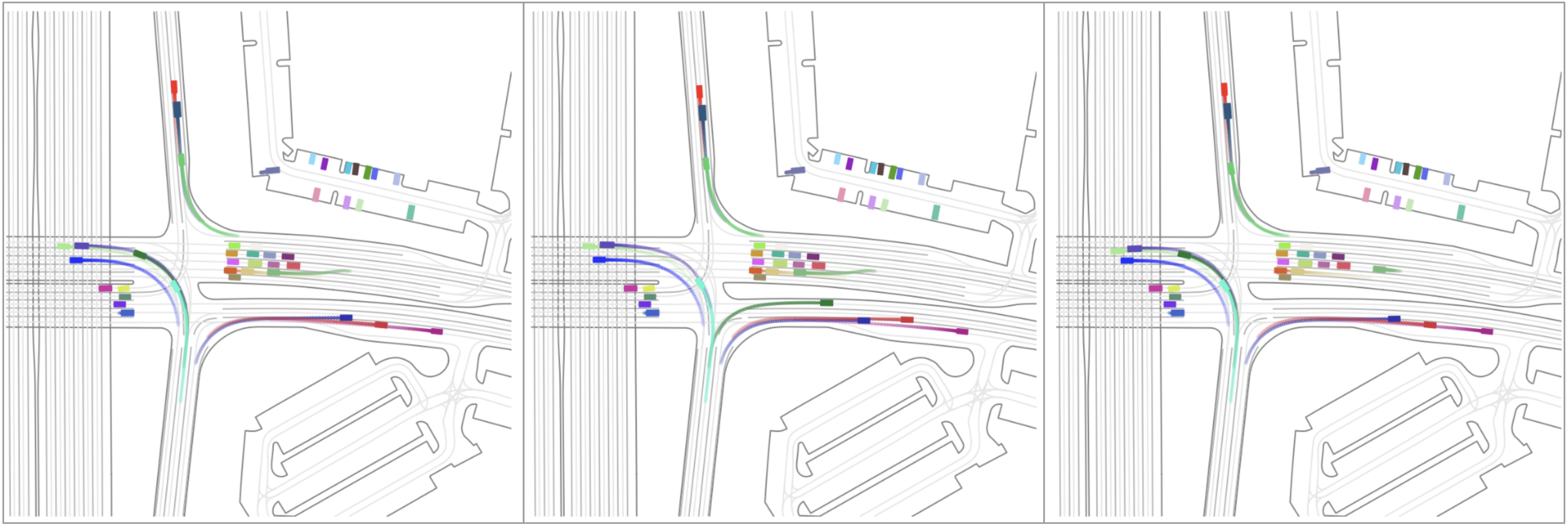}
	\caption{MPS creates diverse (multi-modal)
	rollouts.}
	\label{fig:multi-modal_2}
\end{figure}

\begin{figure}[htb]
	\centering
	\includegraphics[width=\columnwidth]{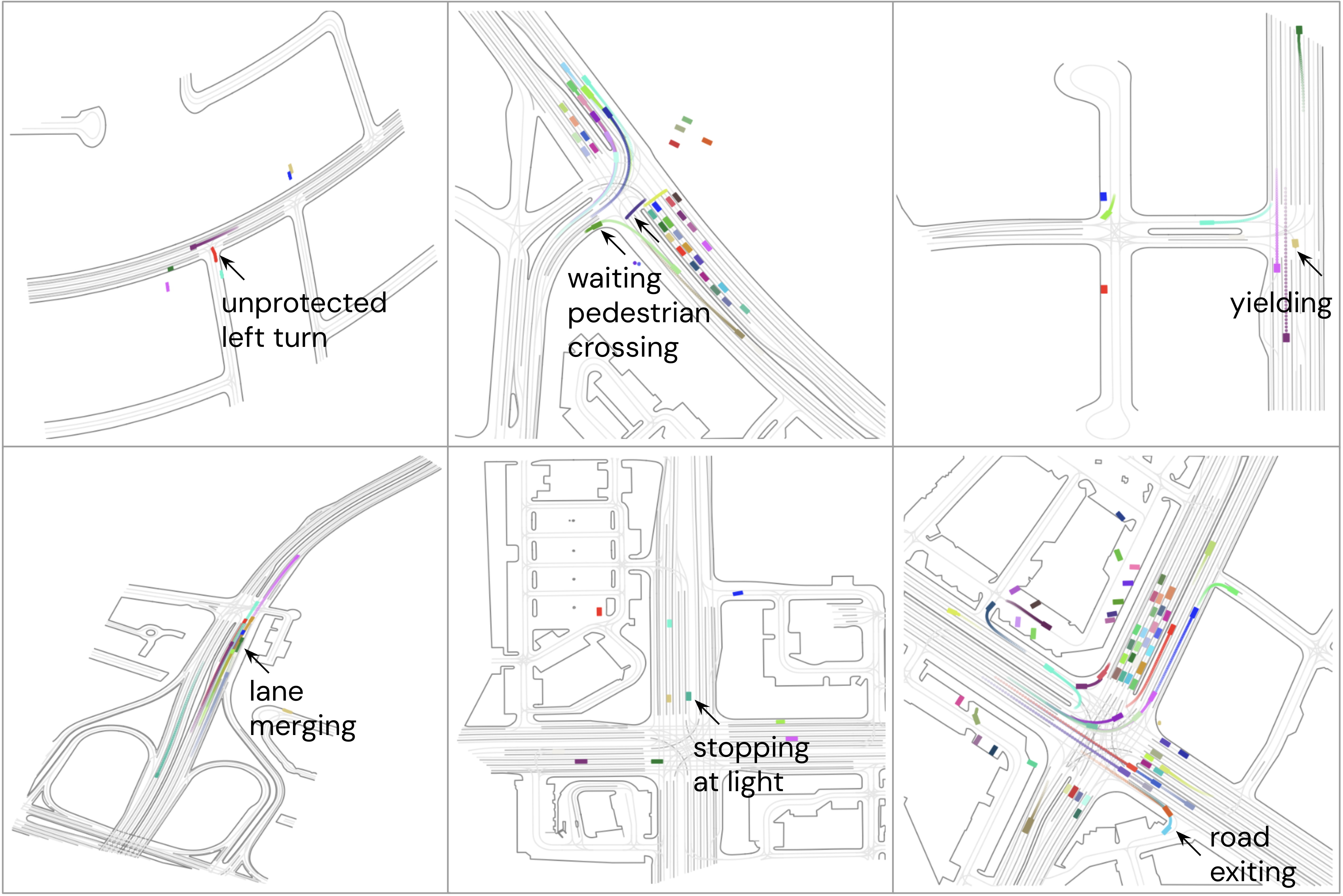}
	\caption{MPS creates realistic traffic patterns.}
	\label{fig:traffic-patterns}
\end{figure}

\noindent
\textbf{Qualitative Study} Qualitatively, MPS generates diverse (multi-modal) predictions (Fig.~\ref{fig:multi-modal_2}),
and each prediction
contains realistic traffic patterns such as lane merging, unprotected left turn, yielding, among others (Fig.~\ref{fig:traffic-patterns}).

%% file: conclusion.tex
\section{Conclusion}

We explored an approach 
that can improve on any trajectory simulation
model by adding domain-specific priors,
and performing inference in the corresponding PGM. We believe combing prior-driven (top-down) and data-driven (bottom-up) methods
is key to building robust and reliable autonomous driving planning and simulation.\footnote{We thank Joseph Ortiz and Wolfgang Lehrach for many useful discussions and suggestions.}

%% file: appendix.tex
\newpage

\section*{Appendix}

\begin{table}[h]
\resizebox{\columnwidth}{!}
{
\begin{tabular}{@{}ccccc@{}}
\toprule
           & mAP    & minADE & minFDE & MissRate \\ \midrule
Vehicle    & 0.4745 & 0.7526 & 1.5107 & 0.1489   \\
Pedestrian & 0.4827 & 0.3455 & 0.7251 & 0.0756   \\
Cyclist    & 0.3898 & 0.7095 & 1.4299 & 0.1865   \\
Avg        & 0.4490 & 0.6025 & 1.2219 & 0.1370   \\ \bottomrule
\end{tabular}
}
\caption{Evaluation of our MTR model on the WOMD Motion Prediction dataset (validation).}
\label{tab:mtr-womd-val}
\end{table}

\begin{figure*}[htb]
	\centering
	\includegraphics[width=\textwidth]{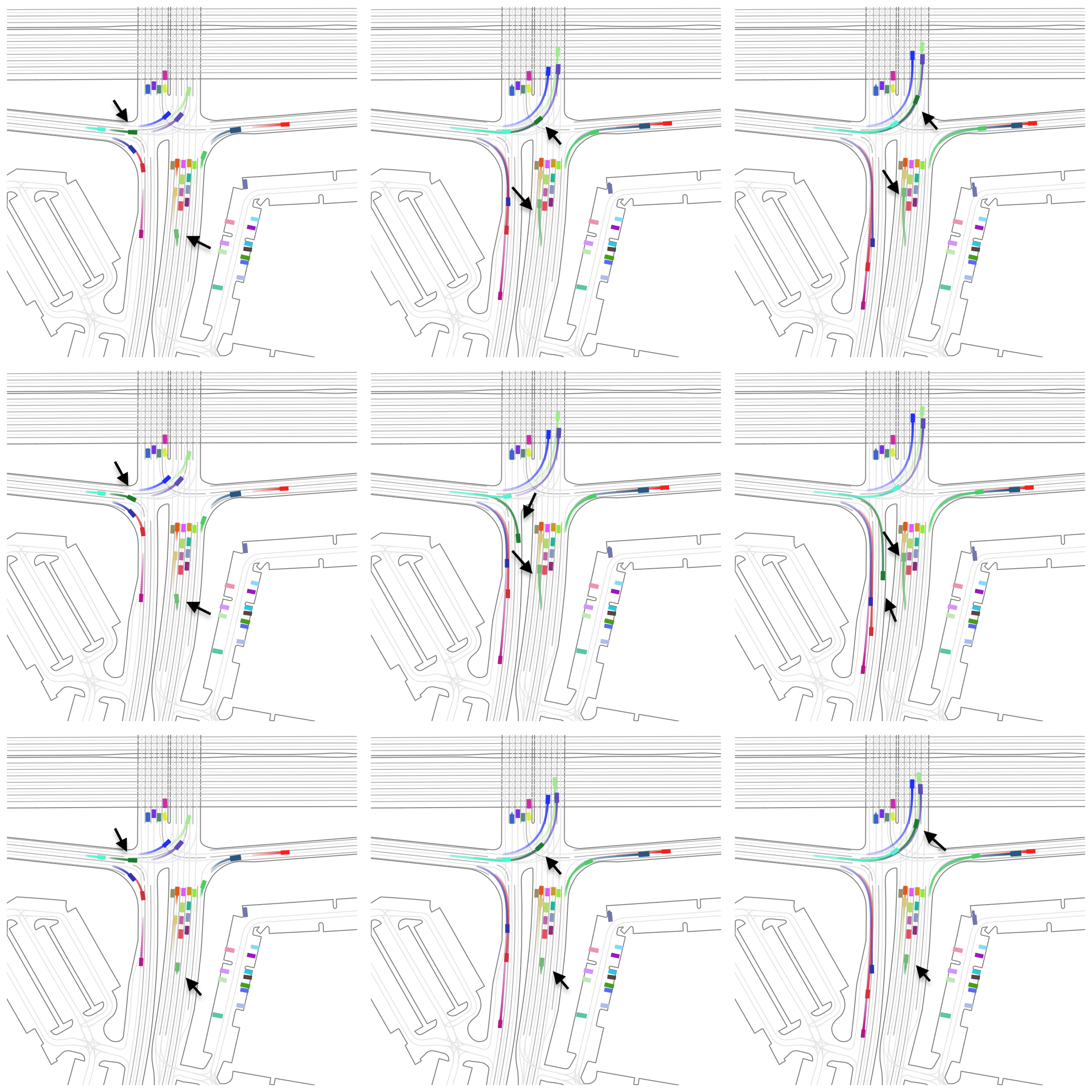}
	\caption{Three simulated scenarios (top to bottom) at different timesteps (left to right) showcasing multi-modal behavior of agents. In the top and bottom simulation, the dark green car takes the left turn. However, in the middle simulation, it turns right. The green car, in the top and middle simulation, attempts the lane change to the left as the cars in front wait at the signal. In the middle simulation, the same green car comes to a stop in the same lane behind the traffic.
	}
	\label{fig:multi-modal}
\end{figure*}

\begin{figure*}[htb]
	\centering
	\includegraphics[width=\textwidth]{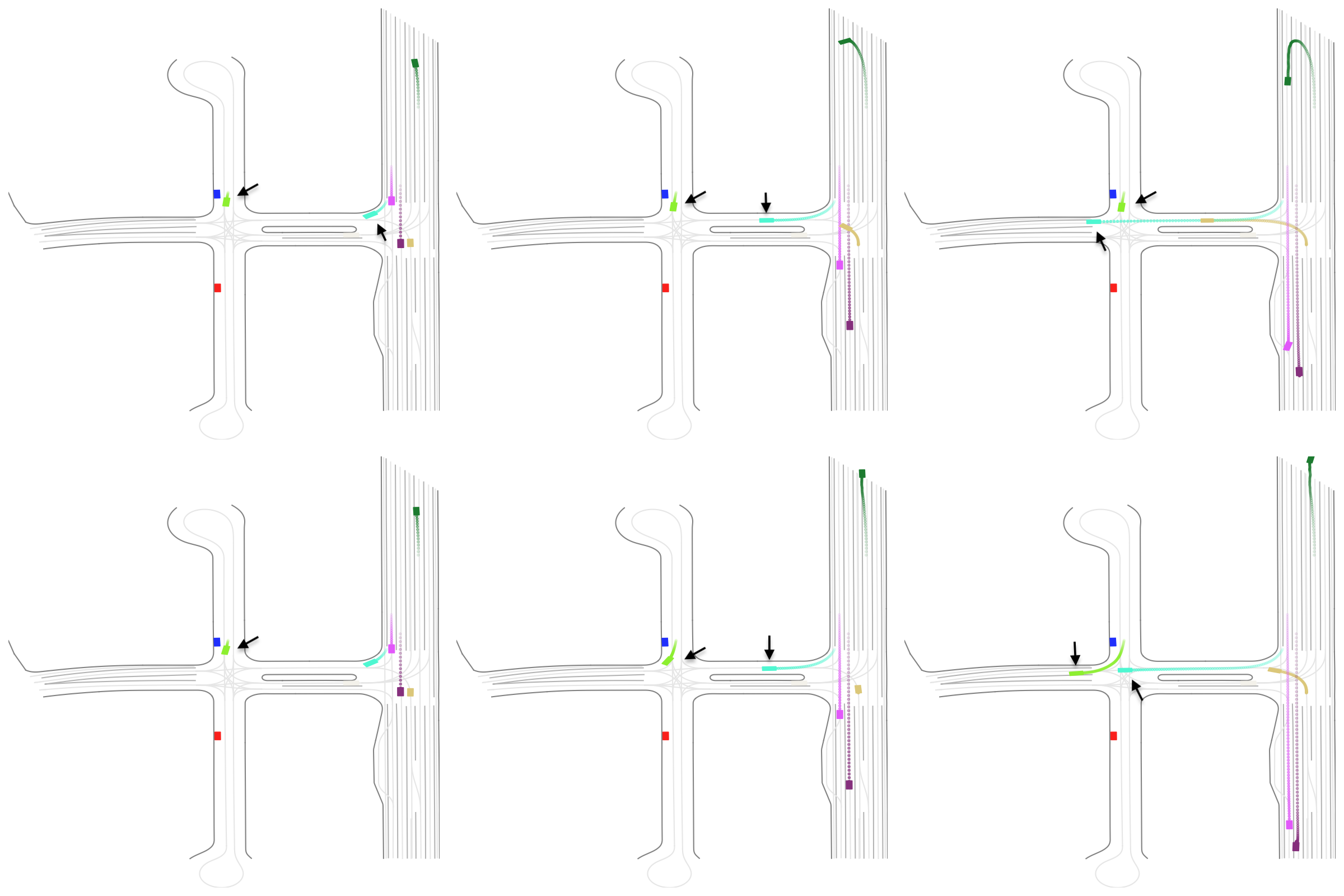}
	\caption{Two simulated scenarios (top to bottom) at different timesteps (left to right). In the top simulation, the light green car, attempting to take the right turn, stops and respects the teal car's right of way. Whereas, in the bottom simulation, the light green car, quickly takes the right turn.
	}
	\label{fig:right-of-way-multi}
\end{figure*}

\begin{figure*}[htb]
	\centering
	\includegraphics[width=\textwidth]{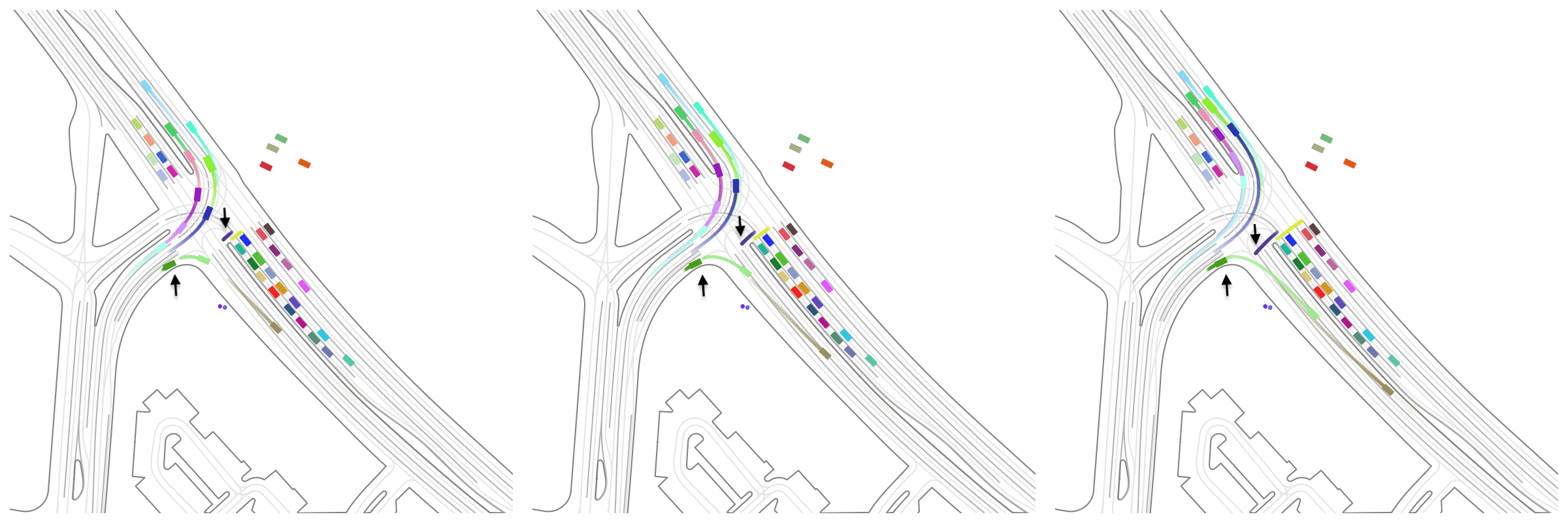}
	\caption{Simulated scenario at different timesteps(left to right). The dark green car, about to take the free right, waits for the pedestrian to cross the road.
	}
	\label{fig:wait-for-pedestrian}
\end{figure*}

\begin{figure*}[htb]
	\centering
	\includegraphics[width=\textwidth]{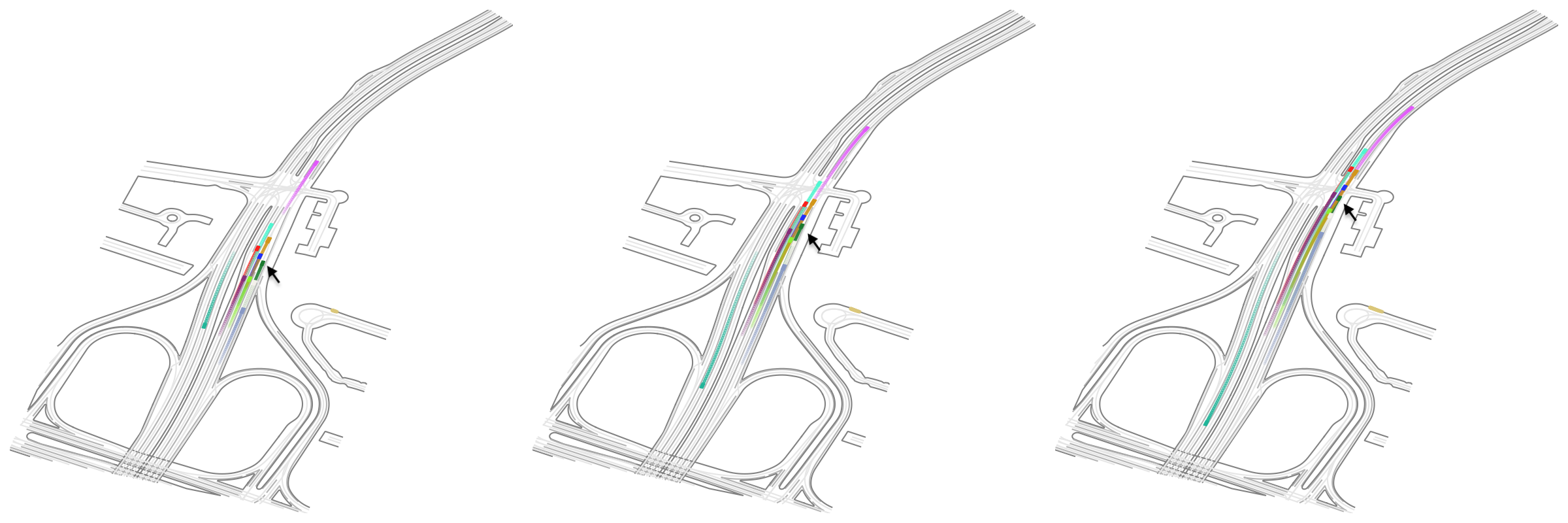}
	\caption{Simulated scenario at different timesteps(left to right). The dark green car merges to the left lane as the right lane comes to an end.
	}
	\label{fig:lane-merge}
\end{figure*}

\begin{figure*}[htb]
	\centering
	\includegraphics[width=\textwidth]{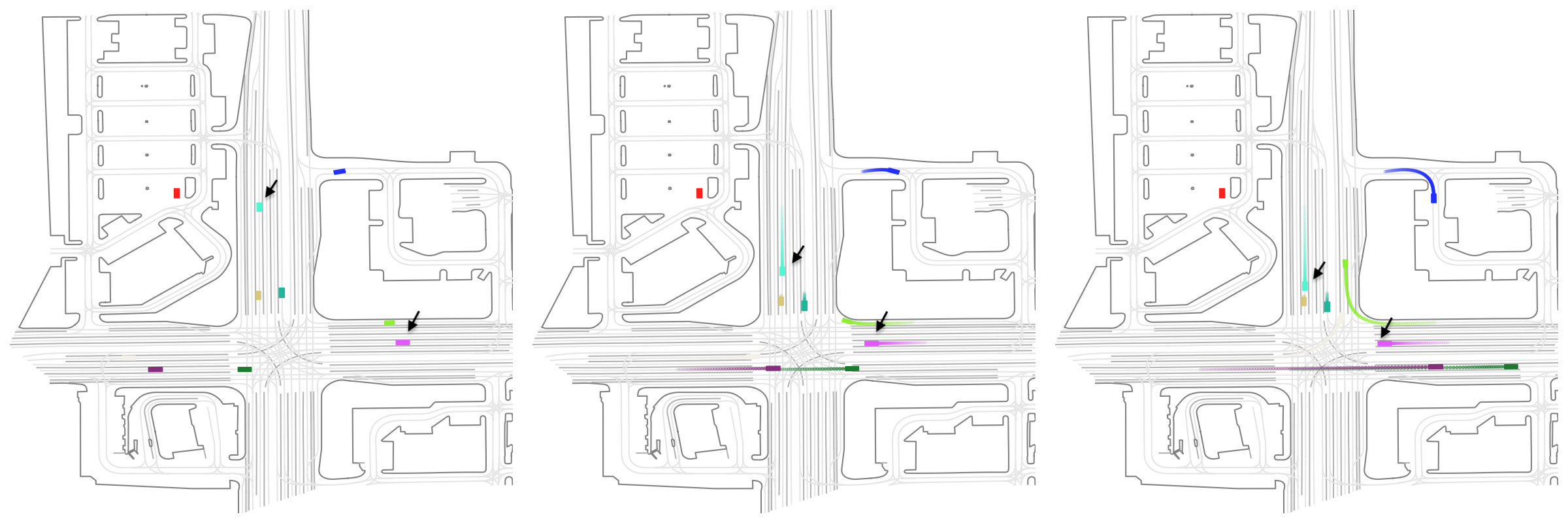}
	\caption{Simulated scenario at different timesteps(left to right). The cars come to stop at the signal. Additionally, the teal car starts to slow down and stops as the golden car in front is waiting at the signal.
	}
	\label{fig:stop-at-signal}
\end{figure*}

\begin{figure*}[htb]
	\centering
	\includegraphics[width=\textwidth]{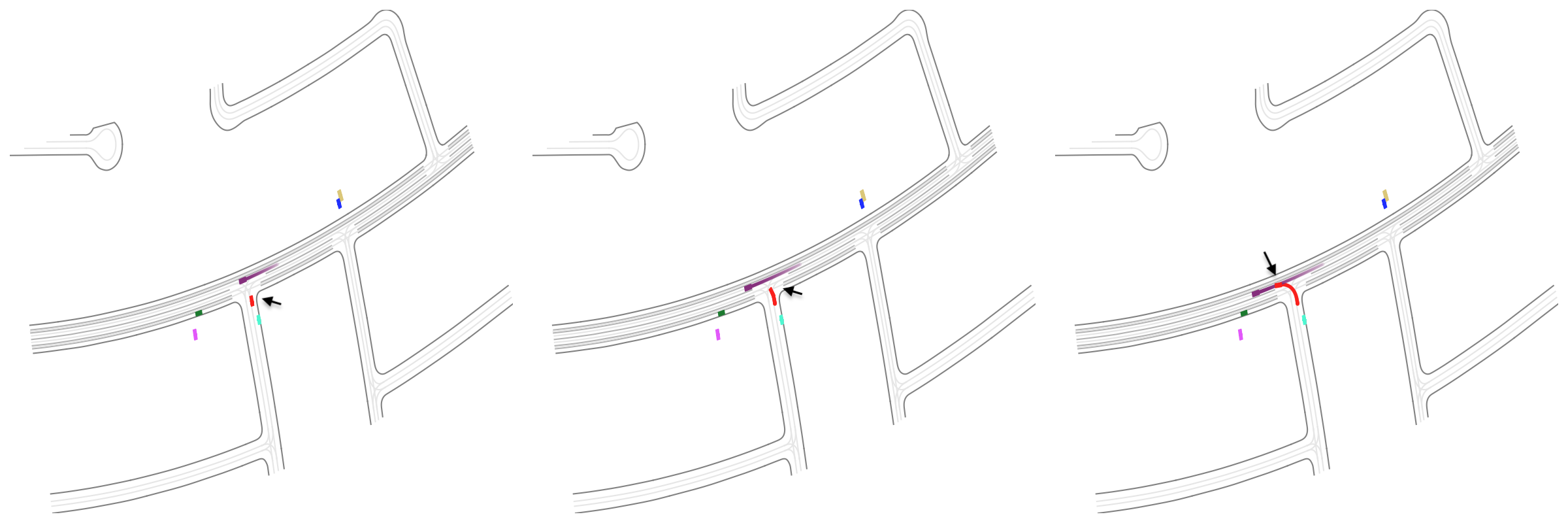}
	\caption{Simulated scenario at different timesteps(left to right). The red car waits for the purple car to pass before taking an unprotected left turn.
	}
	\label{fig:unprotected-left-turn}
\end{figure*}

\begin{figure*}[htb]
	\centering
	\includegraphics[width=\textwidth]{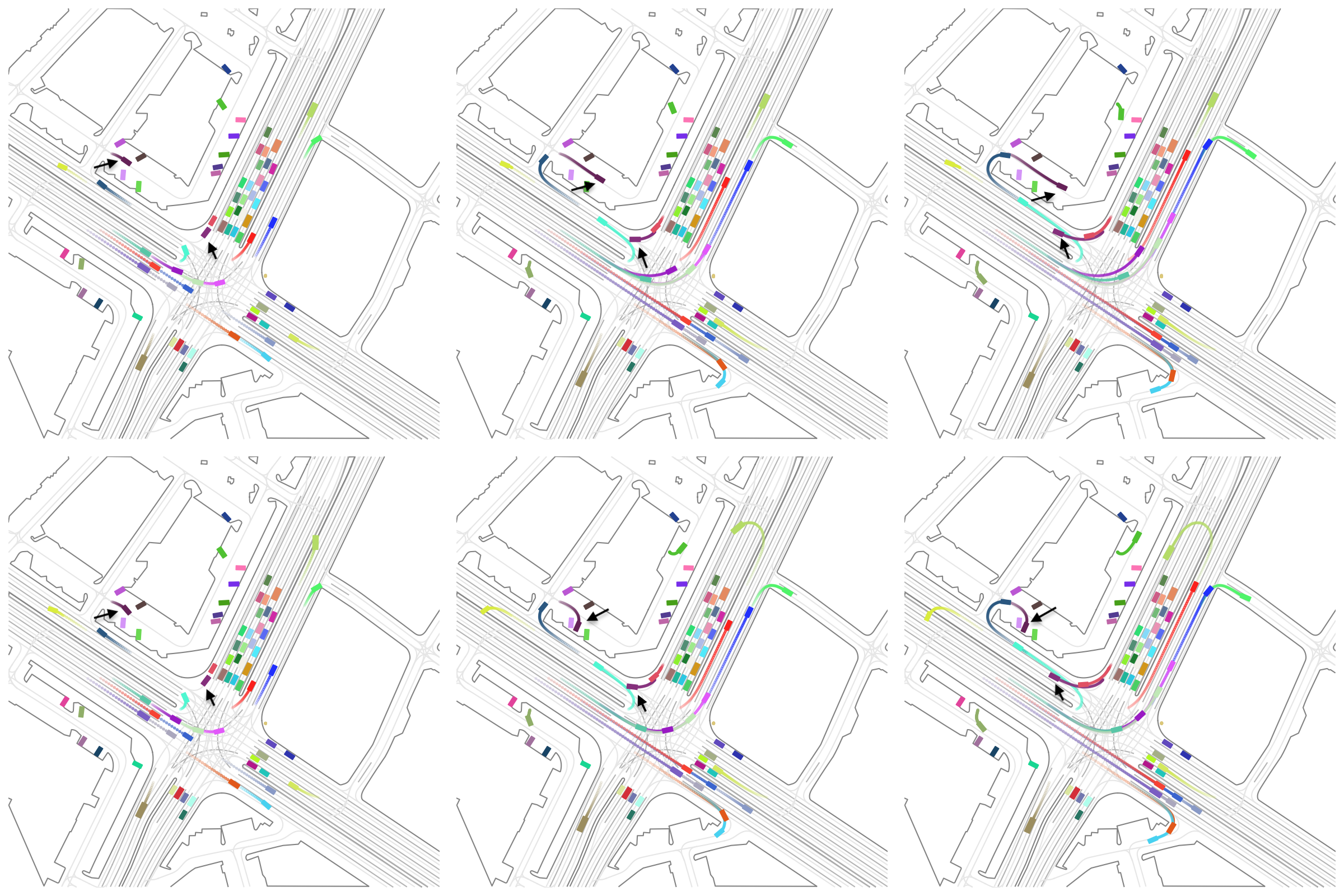}
	\caption{Two simulated scenarios (top to bottom) at different timesteps (left to right). The top simulation shows the brown car entering the parking lot and driving straight. In the bottom simulation, the brown car attempts to park besides the pink car. We can also observe the dark purple car, waiting for the teal car to complete the U-turn before taking the free right.
	}
	\label{fig:car-parks}
\end{figure*}